\documentclass[sigconf,10pt]{acmart}
\renewcommand\footnotetextcopyrightpermission[1]{}

\usepackage{tikz}
\usepackage{amsmath}

\usepackage{enumerate} 
\usepackage{listings}
\usepackage[obeyFinal,textsize=tiny]{todonotes}
\usepackage{verbatim}
\usepackage{paralist}
\usepackage{xspace}
\usepackage{multirow}
\usepackage{subcaption}
\usepackage{booktabs}
\usepackage{adjustbox}
\usepackage[nameinlink,noabbrev]{cleveref}

\graphicspath{{graphics/}}

\AtBeginDocument{%
  }

\setcopyright{acmcopyright}
\copyrightyear{2025}
\acmYear{2025}
\acmDOI{XXXXXXX.XXXXXXX}

\acmConference[GenAIECommerce '25]{Workshop on Generative AI for E-Commerce}{September 22, 2025}{Prague, Czech Republic}
\acmBooktitle{Workshop on Generative AI for E\-Commerce (GenAIECommerce '25),
September 22, 2025, Prague, Czech Republic}
\acmISBN{978-1-4503-XXXX-X/2025/09}

\acmSubmissionID{207}

\newcommand{\assistant}{FaMA\xspace}

\begin{document}

\date{}

\title{\assistant: LLM-Empowered Agentic Assistant for Consumer-to-Consumer Marketplace}

\author{Yineng Yan}
\affiliation{%
  \authornotemark[1]
  \institution{University of Texas at Austin}
  \country{USA}
}
\authornote{Work is done during internship at Meta}
\author{Xidong Wang}
\affiliation{%
  \authornotemark[2]
  \institution{Meta Platforms, Inc}
  \country{USA}
}
\author{Jin Seng Cheng} 
\affiliation{%
  \authornotemark[2]
  \institution{Meta Platforms, Inc}
  \country{USA}
}
\author{Ran Hu}
\affiliation{%
  \authornotemark[2]
  \institution{Meta Platforms, Inc}
  \country{USA}
}
\author{Wentao Guan}
\affiliation{%
  \authornotemark[2]
  \institution{Meta Platforms, Inc}
  \country{USA}
}
\author{Nahid Farahmand}
\affiliation{%
  \authornotemark[2]
  \institution{Meta Platforms, Inc}
  \country{USA}
}
\author{Hengte Lin}
\affiliation{%
  \authornotemark[2]
  \institution{Meta Platforms, Inc}
  \country{USA}
}
\author{Yue Li}
\authornote{corresponding: yueli9@meta.com}
\affiliation{%
  \authornotemark[2]
  \institution{Meta Platforms, Inc}
  \country{USA}
}

\begin{abstract}
The emergence of agentic AI, powered by Large Language Models (LLMs), marks a paradigm shift from reactive generative systems to proactive, goal-oriented autonomous agents capable of sophisticated planning, memory, and tool use. This evolution presents a novel opportunity to address long-standing challenges in complex digital environments. Core tasks on Consumer-to-Consumer (C2C) e-commerce platforms often require users to navigate complex Graphical User Interfaces (GUIs), making the experience time-consuming for both buyers and sellers. This paper introduces a novel approach to simplify these interactions through an LLM-powered agentic assistant. This agent functions as a new, conversational entry point to the marketplace, shifting the primary interaction model from a complex GUI to an intuitive AI agent. By interpreting natural language commands, the agent automates key high-friction workflows. For sellers, this includes simplified updating and renewal of listings, and the ability to send bulk messages. For buyers, the agent facilitates a more efficient product discovery process through conversational search. We present the architecture for Facebook Marketplace Assistant (\assistant), arguing that this agentic, conversational paradigm provides a lightweight and more accessible alternative to traditional app interfaces, allowing users to manage their marketplace activities with greater efficiency. Experiments show \assistant achieves a 98\% task success rate on solving complex tasks on the marketplace and enables up to a 2x speedup on interaction time.
\end{abstract}

\settopmatter{printfolios=true}
\maketitle
\pagestyle{plain}

\section{Introduction}
Consumer-to-Consumer (C2C) e-commerce, with platforms like eBay, Etsy, and Facebook Marketplace, has become a significant part of the digital economy, connecting tens of millions of individual buyers and sellers. However, the user experience on these platforms is often defined by a series of manual, repetitive, and time-consuming tasks that are managed through complex GUIs. This operational friction creates challenges for both sides of the marketplace \cite{cognitive_load}.

For sellers, the workflow involves a sequence of tedious steps. Creating a single, well-optimized product listing can be a lengthy process of data entry and description writing. As their inventory grows, tasks like renewing or relisting items to maintain visibility become a significant time commitment. Furthermore, managing a high volume of inquiries from potential buyers often requires sending similar responses repeatedly, a task that does not scale well for active sellers.

For buyers, the primary challenge lies in product discovery. Navigating a vast and unstructured inventory with basic search and filtering tools can be a frustrating experience. Finding a specific item often requires trying multiple keyword combinations and manually sifting through irrelevant results, making the search process inefficient. These operational hurdles, inherent to the GUI-based model, limit the efficiency and accessibility of C2C platforms for everyday users.

We propose \textbf{Fa}cebook \textbf{M}arketplace \textbf{A}ssistant (\assistant), an LLM-powered agentic assistant \cite{llm_agent} that can serve as a new, conversational entry point to C2C marketplaces, fundamentally simplifying the user experience. By interacting with an agent through natural language, users can delegate complex tasks and bypass the need to navigate the traditional app interface for most core operations. This agent is equipped with a specific set of tools designed to automate the most common and repetitive workflows for both buyers and sellers.

For new users, the agent can facilitate a simpler onboarding process, guiding new participants through community guidelines and helping them create listings via natural language.

For sellers, the agent provides key operations such as listing management and bulk messaging. A seller can simply provide an image and a brief description, and instruct the agent: "Create a new listing for this." The agent can then use its tools to populate the necessary fields and publish the item. Similarly, a command like "Renew all my listings that are expiring this week" automates the otherwise manual renewal process. To manage high-volume communications, a seller can issue a command such as "Reply to all unread messages asking about availability with 'Yes, it's still available'", allowing them to respond to multiple buyers at once.

For buyers, the agent simplifies product discovery. Instead of using rigid filters, a buyer can describe their needs naturally, for example: "I'm looking for a blue vintage jacket in a medium size." The agent interprets this request and uses its search tools to find relevant items, making the discovery process more intuitive and efficient.

This conversational paradigm offers a lightweight and accessible alternative to the traditional app, allowing users to manage their marketplace activities more efficiently.

This paper makes several key contributions to the field of agentic AI for e-commerce. First, we present the complete architecture of \assistant, a conversational agent built on a LLM, a memory system, and a suite of specialized tools for interacting with the marketplace. Second, we evaluate the agent's reliability and reasoning capabilities through an automated framework that shows a 98\% task success rate. We also quantify its real-world efficiency gains through an interaction timing study, which confirms up to a 2x speedup in task completion time. To the best of our knowledge, this work represents the first comprehensive AI agent for C2C e-commerce that provides a unified conversational entry point to the marketplace platform and automates complex tasks for both sellers and buyers.

\section{Related Work}
\paragraph{\textbf{LLM for C2C E-Commerce Platform}}
The application of LLMs in e-commerce has primarily focused on automating specific tasks. IPL \cite{ipl} leverages Multimodal Large Language Models (MLLMs) \cite{mllm} to automatically generate rich, platform-specific product descriptions directly from user-uploaded photos, simplifying one of the most tedious aspects of selling online. FishBargain \cite{fishbargain} is an LLM-empowered agent designed specifically to automate the bargaining process for sellers on C2C e-commerce platforms. It acts as a proactive dialogue agent that understands chat context and product information to negotiate prices with potential buyers, aiming to secure a better deal without the seller's direct intervention.

\paragraph{\textbf{LLM-as-Agent on E-Commerce}}
Beyond single-task automation, agentic frameworks have also been applied to broader domains within e-commerce, such as recommendation. RecMind \cite{RecMind} is a notable example of an agent designed for personalized recommendations. It utilizes an LLM's planning and tool-use capabilities to provide zero-shot recommendations that can adapt to complex user preferences and leverage external knowledge. While powerful within its domain, its focus remains on the product discovery aspect of the user experience.

While these studies highlight the potential of LLMs and AI agents to solve specific challenges, they tend to focus on optimizing a single step in the user workflow. The contribution of \assistant is the design and implementation of a comprehensive, unified AI agent that serves as a holistic assistant for C2C marketplace users, addressing multiple stages of the buyer and seller journeys.

\section{Facebook Marketplace Assistant}
\begin{figure}[h]
\centering
\includegraphics[width=\columnwidth]{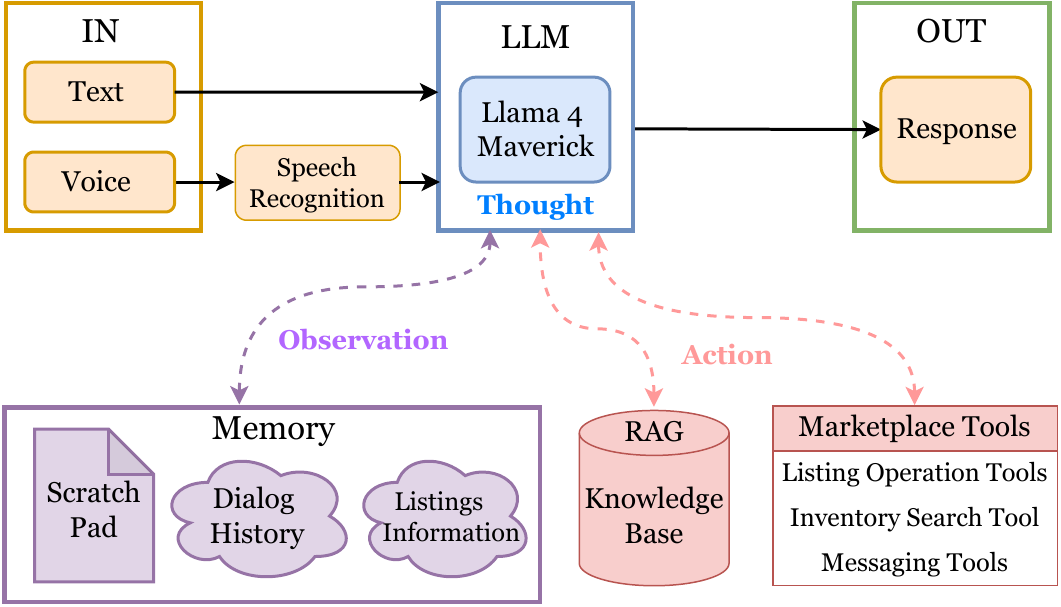}
\caption{Architecture of \assistant.}
\label{fig:arch}
\end{figure}

In this section, we present an overview of the LLM-empowered AI agent architecture for \assistant. As illustrated in \autoref{fig:arch}, the system is composed of three primary components: a Large Language Model (LLM), a memory module, and a suite of tools.

\subsection{Large Language Model}

The Large Language Model (LLM) serves as the core reasoning engine of \assistant. In our implementation, we employ the Llama-4-Maverick-17B-128E-Instruct model \cite{llama4}. This model was specifically chosen for its favorable balance of performance with low computational cost and latency, which are critical considerations for developing a responsive agent suitable for e-commerce applications.

The interaction with the LLM is governed by a carefully structured prompt template. A system prompt initializes the agent, providing it with its persona (e.g., "You are an AI assistant for a Facebook Marketplace user"), a detailed description of its available tools, and instructions on formatting its response.

\begin{figure}[h]
\centering
\includegraphics[width=\columnwidth]{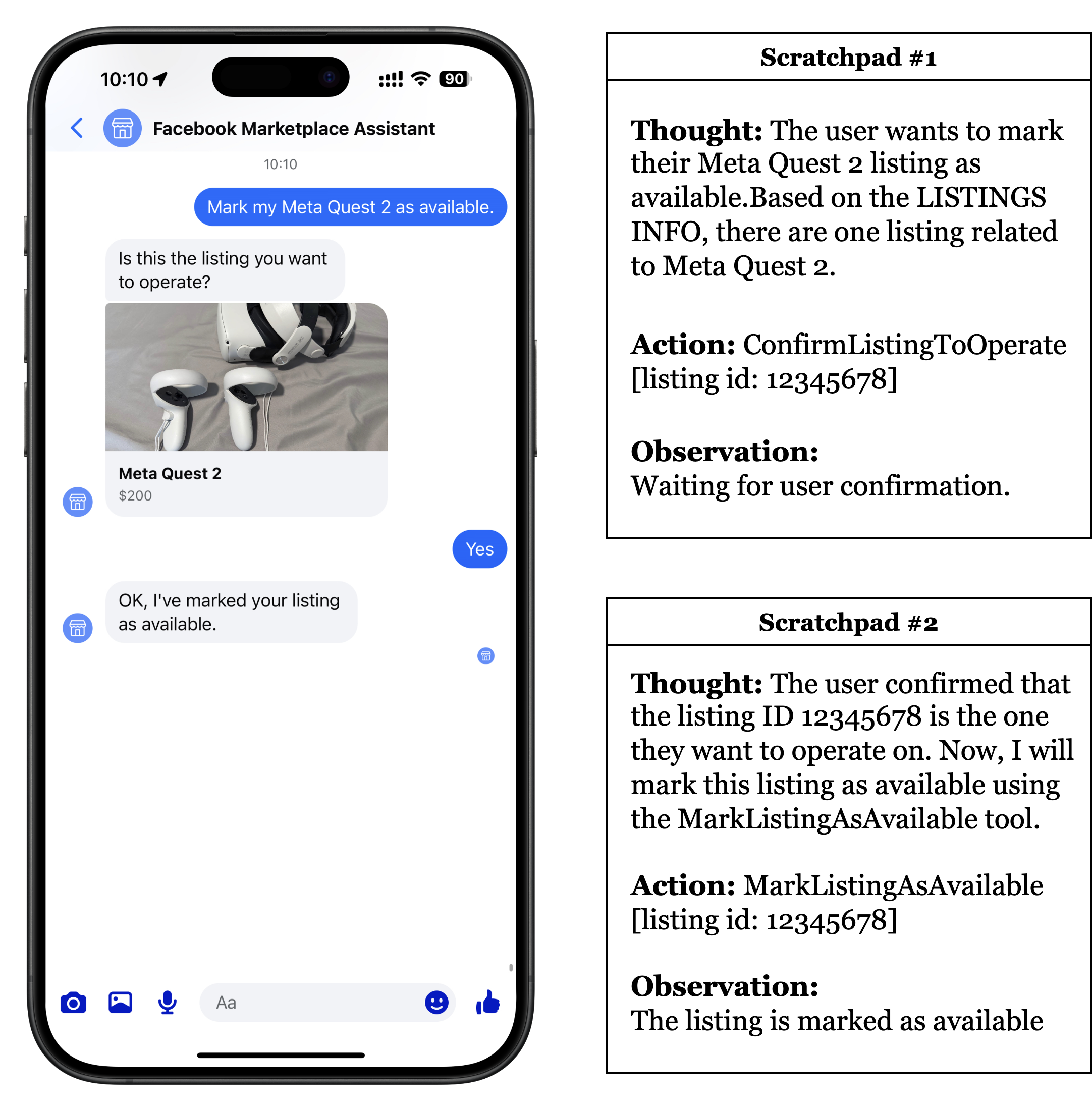}
\caption{The user tells the \assistant which listing to act on by describing the item for sale. The LLM's reasoning process, a sequence of thought-action-observation steps, is saved to a scratchpad shown on the right of the figure.}
\label{fig:scratchpad}
\end{figure}

To effectively synergize reasoning and action, \assistant employs the ReAct (Reasoning and Acting) prompting paradigm \cite{react}. As depicted in \autoref{fig:scratchpad}, the agent solves a task by generating a thought and an action using explicit markers. The \textbf{Thought} is a textual rationale that outlines its current understanding of the task, its progress, and its plan for the next step. The \textbf{Action}, following the thought, is a specific, executable command, which typically manifests as a tool call. The available tools and their usage are detailed in Section \ref{sec:tools}. Within this framework, Chain-of-Thought (CoT) prompting \cite{cot} is utilized to guide the LLM's reasoning process. Once an action is executed, its output is returned to the agent as an \textbf{Observation}. This observation is appended to the context, allowing the LLM to assess the outcome of its previous action and plan its next move.

In a key difference from the standard ReAct agent, where the thought-action-observation cycle runs autonomously until task completion, \assistant operates in a single-step interactive mode. After each cycle, the agent pauses to present its proposed action to the user for confirmation before execution. This design is intentionally tailored to the nature of platforms like Facebook Marketplace for two primary reasons. First, many operations involve critical state changes to a user's account (e.g., sending a message, modifying a listing), making explicit user consent essential for safety and correctness. Second, this interactive approach enhances the user experience by providing transparent, step-by-step updates and allowing for immediate correction, which mitigates long wait times and prevents erroneous actions.

\paragraph{\textbf{Voice Message}} To accommodate users who prefer voice input—a modality often faster and more convenient than typing on mobile devices \cite{speech_messages}—we augmented \assistant with speech processing capabilities. Since the base language model operates exclusively on text, an upstream Automatic Speech Recognition (ASR) \cite{asr} module is integrated into the architecture. This component's sole function is to receive the user's raw audio input and transcribe it into a textual format. The resulting transcript is then passed to the LLM for processing, making it indistinguishable from a standard typed message from the agent's perspective.

\subsection{Memory}
\paragraph{\textbf{Scratchpad}} The single-step interactive design of \assistant, where the agent pauses for user confirmation after each action, introduces a significant challenge in managing conversational state. Since some tasks require multi-step actions, the agent requires a more structured memory mechanism. To maintain context and state across these user interactions, the agent must be able to recall its previous steps. To address this, we implement a memory module based on the "scratchpad" concept \cite{scratchpads}, a technique proven effective for enabling language models to track intermediate steps in complex computations. Our scratchpad functions as a structured, short-term memory buffer. It is implemented as a chronological log of Thought-Action-Observation triplets. At the beginning of a new interactive turn, the complete, ordered history from the scratchpad is retrieved. This historical log is prepended to the user's latest input before being passed to the LLM. By providing this explicit record, the LLM is fully re-contextualized. It understands not only what the user is asking now, but also the preceding reasoning, the actions already taken, and the data it is working with. This enables \assistant to seamlessly continue complex, multi-step dialogues, ensuring goal coherence without repeating work or losing track of the user's ultimate objective. \autoref{fig:scratchpad} shows an example of how the strachpad looks like.
 
\paragraph{\textbf{Dialog History}} \assistant saves the dialogue history that includes user input and agent responses, in memory. This history, however, is intentionally designed to be ephemeral. Each conversation is treated as a distinct session, and the dialogue history is automatically purged after a predefined period of user inactivity. This session-based memory architecture is motivated by three key factors.

First, and most critically, it serves a privacy function. By enforcing a short data retention policy, this design ensures that potentially sensitive user information is not stored long-term. Secondly, it mimics the conventional user experience of human-led support chats, where closing the conversation window typically results in a clean state for the next conversation. This creates a predictable and familiar environment for the user. Finally, from a technical perspective, this design  provides the agent with a clean context. This improves its focus on the immediate task and prevents context from unrelated prior conversations from causing interference, while also significantly reducing token consumption and associated costs.

\paragraph{\textbf{Listings Information}}
Many seller-side operations on the marketplace, such as renewing a listing or lowering the price, require the user to specify a target listing. Graphical selection menus or requiring users to supply exact listing IDs can be cumbersome and disrupt the conversational flow. To create a more fluid and natural user experience, we designed \assistant to understand and resolve ambiguous, text-based references to a user's listings. For example, \autoref{fig:scratchpad} shows an example where a seller can simply state, "Mark my Meta Quest 2 as available." and the agent can identify the correct item without needing a specific ID.

To enable this capability, \assistant utilizes a dedicated Listings Information Memory. At the start of each user session, the agent retrieves key details (e.g., title, description, and ID) for all of the seller's listings from their profile. This information is structured and injected into the LLM's context. When a seller refers to a listing during the chat, the LLM can find the listing that matches the user's description by checking the listings information memory. The listings information contains the unique listing ID, which is then used to execute the appropriate action. This approach significantly reduces user effort and provides a more natural way for them to manage listings.

\subsection{Tools}
\label{sec:tools}
To solve complex tasks like operating a listing on Marketplace, the ability to perform tool calling \cite{tool_calling} is a must for an AI agent. Llama 4, the language model we use, supports this feature \cite{llama_tool}. We developed a set of tools for \assistant to interact with Facebook Marketplace. We categorize those tools into the following types:

\paragraph{\textbf{Listing Operation Tools}}
This suite of tools provides sellers with direct, conversational control over their inventory. These tools allow the agent to create new listings, update the details of existing ones (such as price or description), and renew listings to improve their visibility. To perform updates or renewals on the correct item, the agent intelligently uses the listings information memory to identify the specific listing the user is referring to.

\paragraph{\textbf{Inventory Search Tools}}
This tool serves as the agent's primary interface for buyers to explore the marketplace. It gives the agent the ability to query the marketplace search API based on a user's conversational request. The LLM's role is to parse the user's language, such as "show me used iPhone 13 under \$200," and extract key information like the item name, price constraints, and desired condition. The agent then uses these extracted details to perform a search and present the findings to the user.

\paragraph{\textbf{Messaging Tools}}
Communication is central to e-commerce, and this tool enables the agent to interact with other users on the platform on the user's behalf. Its capabilities include sending an AI-generated message to a user to initiate contact about a specific product. It also supports bulk messaging actions, allowing a buyer to efficiently contact multiple sellers about similar items or a seller to send a standardized response to multiple inquiries at once.

\paragraph{\textbf{RAG as a Tool}}
We treat Retrieval-Augmented Generation (RAG) \cite{rag} as a specialized informational tool. The base LLM lacks deep, up-to-date knowledge of specific marketplace policies or "how-to" guides. To address this, we equipped \assistant with a tool to access a marketplace knowledge base. When the LLM's reasoning determines that a user's question requires platform-specific knowledge, it calls this tool. The tool then initiates a RAG process: it uses the question to retrieve relevant documents from a vector database of help articles and then uses that grounded information to synthesize an accurate, trustworthy answer. This modular design keeps the agent's reasoning separate from the knowledge retrieval mechanism, and avoids unnecessary RAG when a user query does not require that knowledge.

\section{Experiments}

In this section, we present experiments designed to evaluate \assistant across two key dimensions: its reasoning and tool-use accuracy in task completion, and its potential to reduce user effort.

\subsection{Automated Evaluation of Task Performance}

\begin{figure}[h]
\centering
\includegraphics[width=\columnwidth]{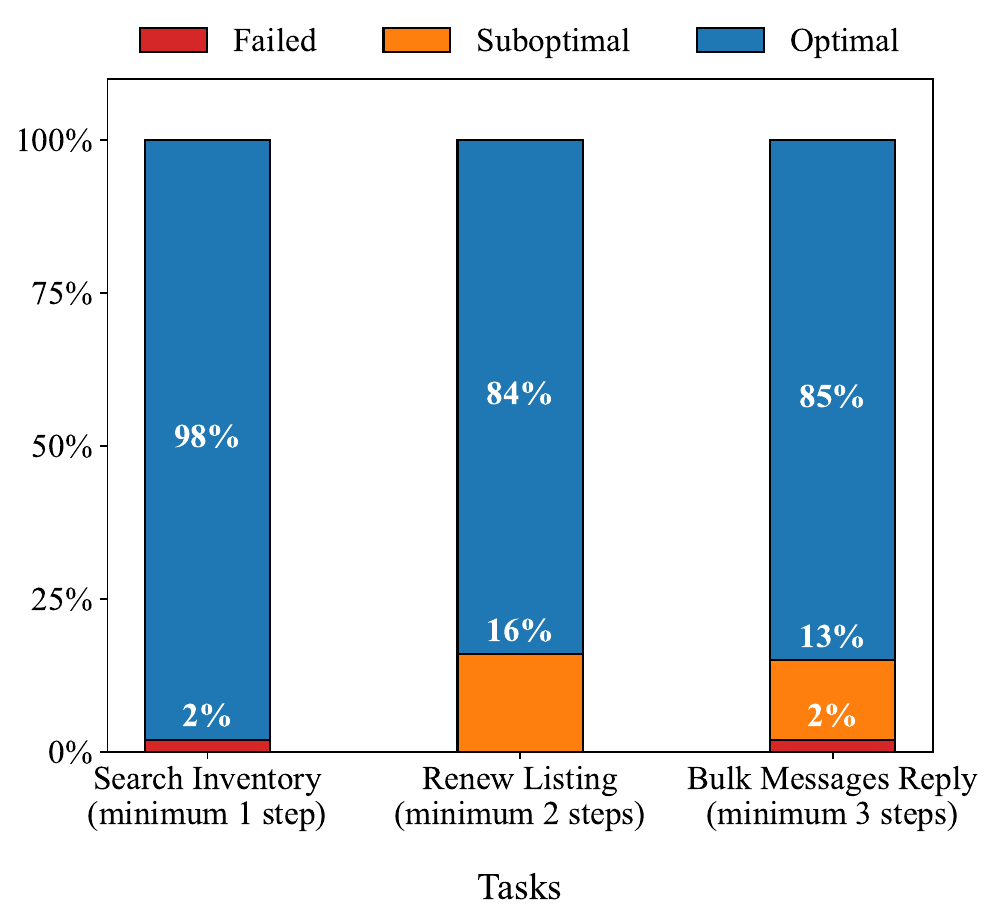}
\caption{Task success and optimality rates for the automated evaluation of \assistant on representative tasks. The overall success rate is 98\% or higher and the optimality rate is above 84\%.}
\label{fig:automation}
\end{figure}

To quantitatively assess the agent's core performance, we developed an automated evaluation pipeline. This pipeline leverages an LLM-based user simulator to interact with \assistant, allowing for scalable and repeatable testing. For consistency and to isolate the agent's architectural merits, both the user simulator and \assistant were powered by the same Llama-4-Maverick-17B-128E-Instruct model. The evaluation was conducted on a synthetic dataset of 100 diverse marketplace listings (including electronics, vehicles, toys, and furniture) that we pre-generated using an LLM.

The user simulator was prompted with the persona of a typical Facebook Marketplace user interacting with the assistant. For each of the 100 listings in our dataset, the simulator was assigned a set of representative tasks, such as renewing an existing item to improve its visibility, searching the inventory for an item on the marketplace, and performing a bulk reply to multiple inquiries about a listing. To ensure that \assistant was solely responsible for task execution, the simulator was instructed to never perform the tasks itself but to insist on the agent's help. To foster diverse and less deterministic conversational paths, the simulator's temperature was set to 1.0.

Our primary metric is the Task Success Rate, defined as the percentage of tasks completed within a 5-step limit. For all successful tasks, we also measure the Task Optimality Rate, which is the percentage of tasks completed in the minimum possible number of steps. The optimal number of steps is predefined based on the task's inherent complexity, ranging from a single step for a basic inventory search, to two steps for renewing a listing (which requires confirming the listing to operate and renewing the listing), and three steps for a bulk reply (which involves finding the listing, fetching unread threads, and then sending replies). A task is marked as a failure if it is not completed within 5 interactive steps.

The results, summarized in \autoref{fig:automation}, demonstrate \assistant's strong performance. For the single-step Inventory Search task, the agent achieved a 98\% success rate, with 100\% of successful attempts completed in the optimal single step. This indicates a high proficiency in basic tool selection and execution. For the more complex, multi-step tasks of Renew Listing and Bulk Reply, \assistant also performed robustly, achieving a success rate of over 96\%. Among these successful multi-step tasks, more than 84\% were completed in the optimal number of steps. The overall failure rate across these complex tasks remained below 2\%. This suggests that the agent's reasoning capabilities and its ability to maintain state and context using the scratchpad memory, are effective for orchestrating multi-tool workflows. The 100\% success rate on the Renew Listing task also indicates that the agent can effectively distinguish the correct item from the 100 listings stored in the Listings Information Memory based on a user's natural language description.

\subsection{Reduced User Effort}
\begin{table}[h]
\begin{center}
 \caption{Comparison of the interaction time to complete a task using \assistant versus manual operation on the Marketplace app on a mobile device.}
\begin{tabular}{ |c||c|c|c|  }
 \hline
 \multicolumn{4}{|c|}{Interaction Time} \\
 \hline
 Task & \assistant & Manual & Speedup\\
 \hline
 Bulk Messages Reply & 25 sec & 50 sec & 2x\\
 Inventory Search & 15 sec & 25 sec & 1.66x\\
 \hline
\end{tabular}
\label{tab:time}
\end{center}
\end{table}

Many routine tasks on e-commerce platforms can be time-consuming and cumbersome. For sellers, popular listings can generate numerous inquiries that require repetitive, individual replies. For buyers, navigating complex interfaces with multiple filters to find a specific item can be a non-trivial task. This experiment aims to quantify the efficiency gains offered by \assistant for these types of workflows.

To measure this, we conducted a comparative timing analysis. As experienced users of the Facebook Marketplace, we timed the process of completing two common tasks: once manually using the standard mobile application, and once conversationally using \assistant. The two evaluated tasks were: (1) Bulk Messaging, defined as sending a reply to five unread inquiries for a single listing, and (2) Filtered Inventory Search, defined as searching for an item with three specific filters applied (e.g., condition, price, and location).

The results of this analysis are presented in \autoref{tab:time}. \assistant achieved a speedup of up to 2x. These findings indicate that \assistant can substantially reduce the time and interactive steps required for common, multi-part tasks, directly demonstrating its value in reducing user effort.

\section{Conclusion}
In this paper, we introduced the Facebook Marketplace Assistant (\assistant), an LLM-powered agentic assistant designed for C2C e-commerce platforms. We presented a comprehensive architecture that integrates a Large Language Model for reasoning, a memory system to maintain state and context, and a suite of specialized tools to interact directly with the marketplace platform. Our experiments show that FMA is highly effective, achieving a 98\% success rate in complex, automated evaluations and enabling up to a 2x speedup in common user workflows compared to manual operation.

\bibliographystyle{ACM-Reference-Format}

\begin{thebibliography}{10}

\bibitem{fishbargain}
Dexin Kong, Xu Yan, Ming Chen, Shuguang Han, Jufeng Chen, and Fei Huang.
\newblock {FishBargain: An LLM-Empowered Bargaining Agent for Online Fleamarket Platform Sellers}.
\newblock In {\em Companion Proceedings of the ACM on Web Conference 2025}, WWW '25, pages 2855--2858, New York, NY, USA, 2025. Association for Computing Machinery.

\bibitem{ipl}
Kang Chen, Qing~Heng Zhang, Chengbao Lian, Yixin Ji, Xuwei Liu, Shuguang Han, Guoqiang Wu, Fei Huang, and Jufeng Chen.
\newblock {IPL}: Leveraging multimodal large language models for intelligent product listing.
\newblock In Franck Dernoncourt, Daniel Preo{\c{t}}iuc-Pietro, and Anastasia Shimorina, editors, {\em Proceedings of the 2024 Conference on Empirical Methods in Natural Language Processing: Industry Track}, pages 697--711, Miami, Florida, US, November 2024. Association for Computational Linguistics.

\bibitem{RecMind}
Yancheng Wang, Ziyan Jiang, Zheng Chen, Fan Yang, Yingxue Zhou, Eunah Cho, Xing Fan, Yanbin Lu, Xiaojiang Huang, and Yingzhen Yang.
\newblock {RecMind}: Large language model powered agent for recommendation.
\newblock In Kevin Duh, Helena Gomez, and Steven Bethard, editors, {\em Findings of the Association for Computational Linguistics: NAACL 2024}, pages 4351--4364, Mexico City, Mexico, June 2024. Association for Computational Linguistics.

\bibitem{react}
Shunyu Yao, Jeffrey Zhao, Dian Yu, Nan Du, Izhak Shafran, Karthik Narasimhan, and Yuan Cao.
\newblock React: Synergizing reasoning and acting in language models.
\newblock In {\em International Conference on Learning Representations (ICLR)}, 2023.

\bibitem{scratchpads}
Maxwell Nye, Anders~Johan Andreassen, Guy Gur-Ari, Henryk Michalewski, Jacob Austin, David Bieber, David Dohan, Aitor Lewkowycz, Maarten Bosma, David Luan, and others.
\newblock Show your work: Scratchpads for intermediate computation with language models.
\newblock 2021.

\bibitem{tool_calling}
Timo Schick, Jane Dwivedi-Yu, Roberto Dessi, Roberta Raileanu, Maria Lomeli, Eric Hambro, Luke Zettlemoyer, Nicola Cancedda, and Thomas Scialom.
\newblock Toolformer: Language models can teach themselves to use tools.
\newblock In {\em Thirty-seventh Conference on Neural Information Processing Systems}, 2023.

\bibitem{llama_tool}
Meta.
\newblock Tool calling with llama: Enhancing ai capabilities.
\newblock {\em Llama Cookbook}, 2025.

\bibitem{speech_messages}
Sherry Ruan, Jacob~O Wobbrock, Kenny Liou, Andrew Ng, and James~A Landay.
\newblock Comparing speech and keyboard text entry for short messages in two languages on touchscreen phones.
\newblock {\em Proceedings of the ACM on Interactive, Mobile, Wearable and Ubiquitous Technologies}, 1(4):1--23, 2018.

\bibitem{llama4}
Meta.
\newblock Llama-4-maverick-17b-128e-instruct, 2025.

\bibitem{cot}
Jason Wei, Xuezhi Wang, Dale Schuurmans, Maarten Bosma, Fei Xia, Ed Chi, Quoc~V Le, Denny Zhou, and others.
\newblock Chain-of-thought prompting elicits reasoning in large language models.
\newblock {\em Advances in neural information processing systems}, 35:24824--24837, 2022.

\bibitem{asr}
Dong Yu and Lin Deng.
\newblock {\em Automatic speech recognition}, volume~1.
\newblock Springer, 2016.

\bibitem{rag}
Patrick Lewis, Ethan Perez, Aleksandra Piktus, Fabio Petroni, Vladimir Karpukhin, Naman Goyal, Heinrich K{\"u}ttler, Mike Lewis, Wen-tau Yih, Tim Rockt{\"a}schel, and others.
\newblock Retrieval-augmented generation for knowledge-intensive nlp tasks.
\newblock {\em Advances in neural information processing systems}, 33:9459--9474, 2020.

\bibitem{llm_agent}
Zhiheng Xi, Wenxiang Chen, Xin Guo, Wei He, Yiwen Ding, Boyang Hong, Ming Zhang, Junzhe Wang, Senjie Jin, Enyu Zhou, and others.
\newblock The rise and potential of large language model based agents: A survey.
\newblock {\em Science China Information Sciences}, 68(2):121101, 2025.

\bibitem{cognitive_load}
Paul Jen-Hwa Hu, Han-fen Hu, and Xiao Fang.
\newblock Examining the mediating roles of cognitive load and performance outcomes in user satisfaction with a website.
\newblock {\em Mis Quarterly}, 41(3):975--A11, 2017.

\bibitem{mllm}
Shukang Yin, Chaoyou Fu, Sirui Zhao, Ke Li, Xing Sun, Tong Xu, and Enhong Chen.
\newblock A survey on multimodal large language models.
\newblock {\em National Science Review}, 11(12):nwae403, 2024.

\end{thebibliography}

\end{document}